\documentclass{article}

\PassOptionsToPackage{numbers, compress, sort}{natbib}

 \usepackage[final]{neurips_2020}

\usepackage[utf8]{inputenc} 
\usepackage[T1]{fontenc}    
\usepackage[hyphens]{url}            
\usepackage[hidelinks]{hyperref}       
\usepackage{booktabs}       
\usepackage{amsfonts}       
\usepackage{nicefrac}       
\usepackage{microtype}      
\usepackage{commath}
\usepackage{longtable}
\usepackage[graphicx]{realboxes}
\usepackage{placeins}
\usepackage[font={small}]{caption}
\usepackage{textcomp}
\usepackage{dcolumn}
\usepackage{mathtools}
\usepackage[dvipsnames]{xcolor}
\usepackage{algorithm}
\usepackage{stmaryrd}
\usepackage{algorithmic}
\usepackage{subcaption}
\usepackage{float}
\usepackage[load-configurations=version-1]{siunitx} 
\usepackage{xcolor}

\definecolor{lightblue}{rgb}{0.01, 0.6, 1.0}

\newsavebox\CBox
\def\textBF#1{\sbox\CBox{#1}\resizebox{\wd\CBox}{\ht\CBox}{\textbf{#1}}}

\title{Deep Transfer Learning for Automated Diagnosis of Skin Lesions from Photographs}

\author{%
  Doyoon Kim$^*$\\
  Cleveland High School\\
  California, US\\
  \And
  Emma Rocheteau\dag \thanks{Equal contribution, \dag Corresponding author}\\
  Department of Computer Science and Technology\\
  University of Cambridge, UK\\
  \texttt{ecr38@cam.ac.uk}\\
}

\begin{document}

\maketitle

\begin{abstract}

Melanoma is not the most common form of skin cancer, but it is the most deadly. Currently, the disease is diagnosed by expert dermatologists, which is costly and requires timely access to medical treatment. Recent advances in deep learning have the potential to improve diagnostic performance, expedite urgent referrals and reduce burden on clinicians. Through smart phones, the technology could reach people who would not normally have access to such healthcare services, e.g. in remote parts of the world, due to financial constraints or in 2020, COVID-19 cancellations. To this end, we have investigated various transfer learning approaches by leveraging model parameters pre-trained on ImageNet with finetuning on melanoma detection. We compare EfficientNet, MnasNet, MobileNet, DenseNet, SqueezeNet, ShuffleNet, GoogleNet, ResNet, ResNeXt, VGG and a simple CNN with and without transfer learning. We find the mobile network, EfficientNet (with transfer learning) achieves the best mean performance with an area under the receiver operating characteristic curve (AUROC) of 0.931$\pm$0.005 and an area under the precision recall curve (AUPRC) of 0.840$\pm$0.010. This is significantly better than general practitioners (0.83$\pm$0.03 AUROC) and dermatologists (0.91$\pm$0.02 AUROC).

\end{abstract}

\section{Introduction}
Melanoma is the most common cause of skin cancer related deaths worldwide \cite{10.1093/jjco/hyy162}. In the United States alone, it is estimated that there will be 100,350 cases and 6,850 melanoma-related deaths in 2020~\cite{americancancersociety}. Initially, it develops in melanocytes where genetic mutations lead to unregulated growth and the ability to metastasise to other areas of the body~\cite{Zbytek2008}. Like many cancers, early detection is key to successful treatment. If melanoma is detected before spreading to the lymph nodes, the average five-year survival rate is 98\%. However, this drops to 64\% if it has spread to regional lymph nodes, and 23\% if it has reached distant organs.

Currently, melanoma is diagnosed by professional medical examination \cite{Kittler}. A meta-analysis conducted by \citet{Phillips2019} showed that when distinguishing between melanoma and benign skin lesions, primary care physicians (10 studies) achieve an area under the receiver operating characteristic curve (AUROC) of 0.83$\pm$0.03, and dermatologists (92 studies) achieve 0.91$\pm$0.02. 

Recent advances in deep learning have the potential to improve diagnostic performance and improve access to care for those in need \cite{Fujisawa}. Through mobile application technology, it will soon be possible to diagnose, refer and provide follow-ups to patients in the community. However, worrying findings from \citet{smartphoneapps} suggest the majority (if not all) diagnostic applications on app stores have not been clinically validated. As a result, clinicians are becoming increasingly concerned about misleading applications \cite{misleadingapps}. However, we believe that when models are developed collaboratively with clinicians, and when they are rigorously evaluated with significance testing and model interpretability, useful tools can be produced to support health in the community.

In this work, we investigate transfer learning with various Convolutional Neural Networks (CNNs) on the binary classification task of classifying melanoma and benign skin lesions. In addition, we perform post-hoc visualisation of the feature attributions using integrated gradients \cite{integratedgradients}.

\section{Related Work}
Recent work by \citet{NIPS2019_8596} cast doubt on the usefulness of transfer learning (TL) for medical imaging. However, a few TL works have achieved success on the problem of melanoma detection~\cite{zunair, 7792699, DBLP:journals/corr/MurphreeN17, romero, 10.1007/978-981-13-7564-4_32, 10.1371/journal.pone.0217293} (although they do not necessarily compare the model with and without TL). We did not find an extensive survey on existing TL models such as ours. The highest AUROC for melanoma detection that we found on photographs was 0.880 in~\citet{DBLP:journals/corr/abs-1902-06061}, which is still lower than the performance for professional dermatologists found in \citet{Phillips2019}.

\section{Methods}
\label{methods}
\begin{figure}[h]
  \centering
  \includegraphics[width=0.93\textwidth]{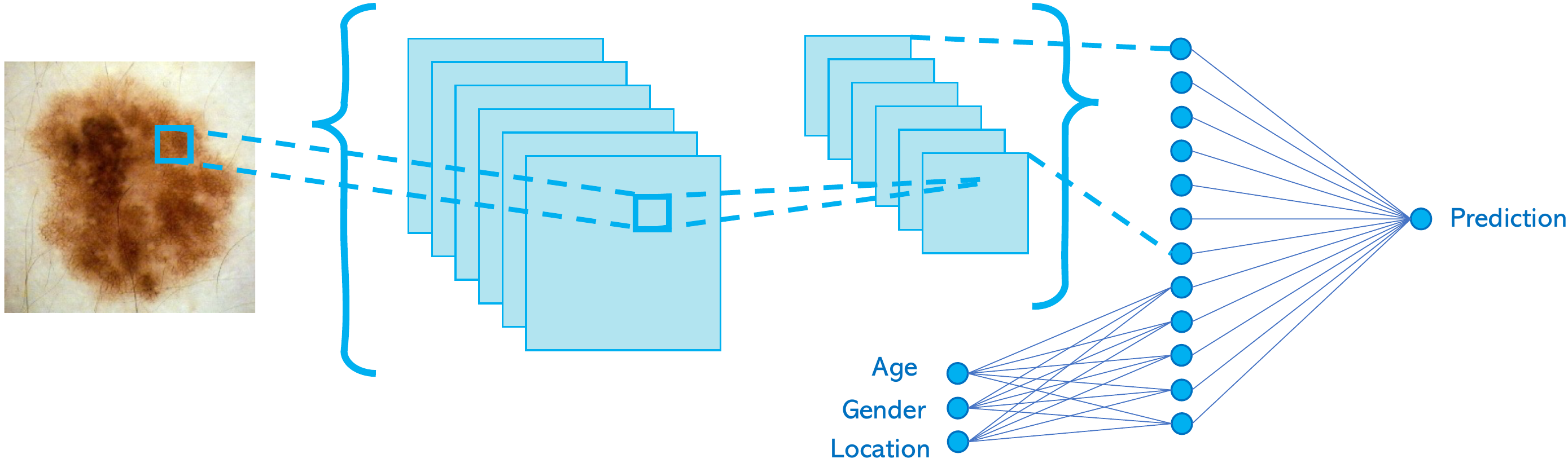}
  \caption{Model architecture. The CNN component (indicated in brackets) is different in each experiment. The static data is processed separately and concatenated to the CNN output before a final prediction is made.}
\label{fig:modelarchitecture}
\end{figure}

Our task is to classify between benign nevi and malignant melanomas. For each patient we have a dermatoscopic photograph in RBG format, $\mathbf{x}\in\mathbb{R}^{3\times224\times224}$ and static features, $\mathbf{s}\in\mathbb{R}^{3}$ (age, gender and location on the body) and the binary label $y\in\mathbb{R}^{1}$. Figure~\ref{fig:modelarchitecture} shows the basic architecture of all models. Our code is publicly available at \url{https://github.com/aimadeus/Transfer_learning_melanoma}.

\subsection{Transfer Learning}
Transfer Learning (TL) is a machine learning method where the weights of a trained model are used to initialise another model on a different task \cite{transferlearning}. In our case, we investigate several CNN architectures using pre-training on ImageNet \cite{imagenet} (a database containing over 14 million images). The last fully connected layer is replaced with one that has a binary output, and whose weights are initialised using Kaiming initialisation \cite{DBLP:journals/corr/HeZR015}. Further description of the various CNN architectures are provided in Appendix~\ref{transfermodels}. We also train a standard 5-layer CNN with no transfer learning (hyperparameter optimisation and further implementation details are provided in Appendix~\ref{hyperparametertuning} and \ref{implementation} respectively).


\section{Experiments}

\subsection{Data}
\begin{figure}[h]
  \centering
      \includegraphics[width=0.2\textwidth]{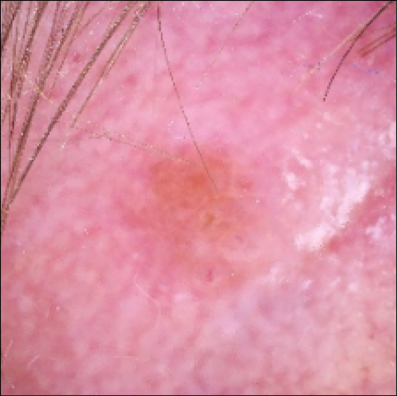}
      \hspace{0.1em}
      \includegraphics[width=0.2\textwidth]{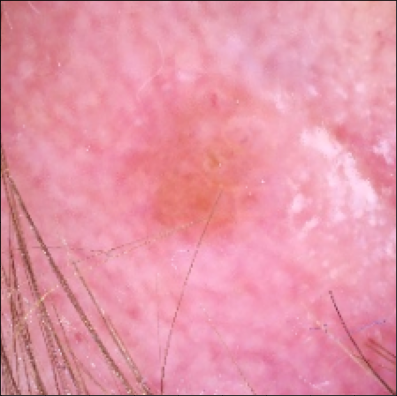}
      \hspace{1em}
      \includegraphics[width=0.2\textwidth]{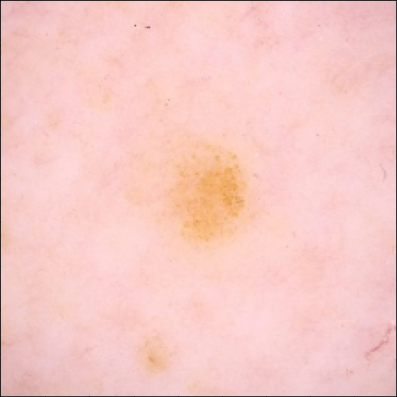}
      \hspace{0.1em}
      \includegraphics[width=0.2\textwidth]{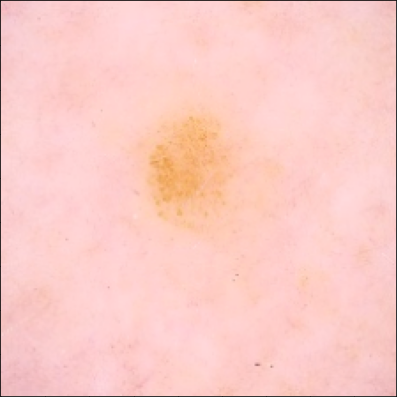}
    \caption{Example photographs in the training data. In each pair of images, the raw data is shown on the left and an augmented image example is shown on the right.}
    \label{augmented}
\end{figure}

We use the International Skin Imaging Collaboration (ISIC) 2020 dataset~\cite{dataset1} (released August 2020), containing labelled photographs taken from various locations on the body (see Table~\ref{image_distribution} in the Appendix). We noted a significant class imbalance with only 2\% of the data containing melanoma. To improve this ratio, we added a second dataset with additional malignant cases \cite{dataset2}, which brought the total to 37,648 skin lesion images. The data was split such that 60\%, 20\% and 20\% was used for training, validating and testing respectively. Data Augmentation was performed on the training data to introduce small variations in the form of random rotations, horizontal and vertical flipping, resizing, brightness, and saturation shifts. This means the training data is subtly altered each time it is presented to the model. Figure~\ref{augmented} shows two examples of raw and augmented images respectively.

\subsection{Results}
\begin{figure}[h]
\centering
\begin{minipage}{.56\textwidth}
  \centering
  \includegraphics[width=.492\linewidth]{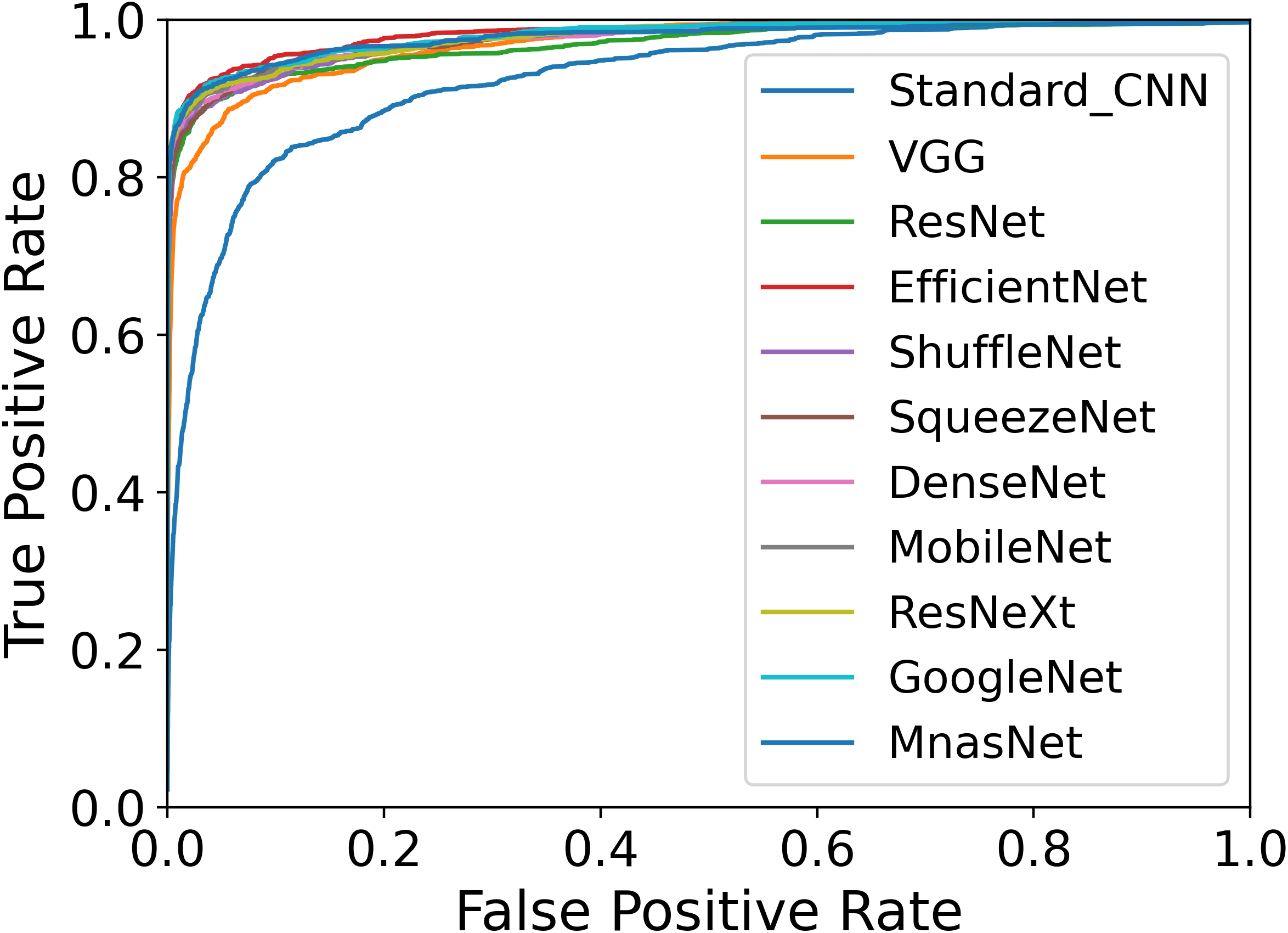}
  \includegraphics[width=.492\linewidth]{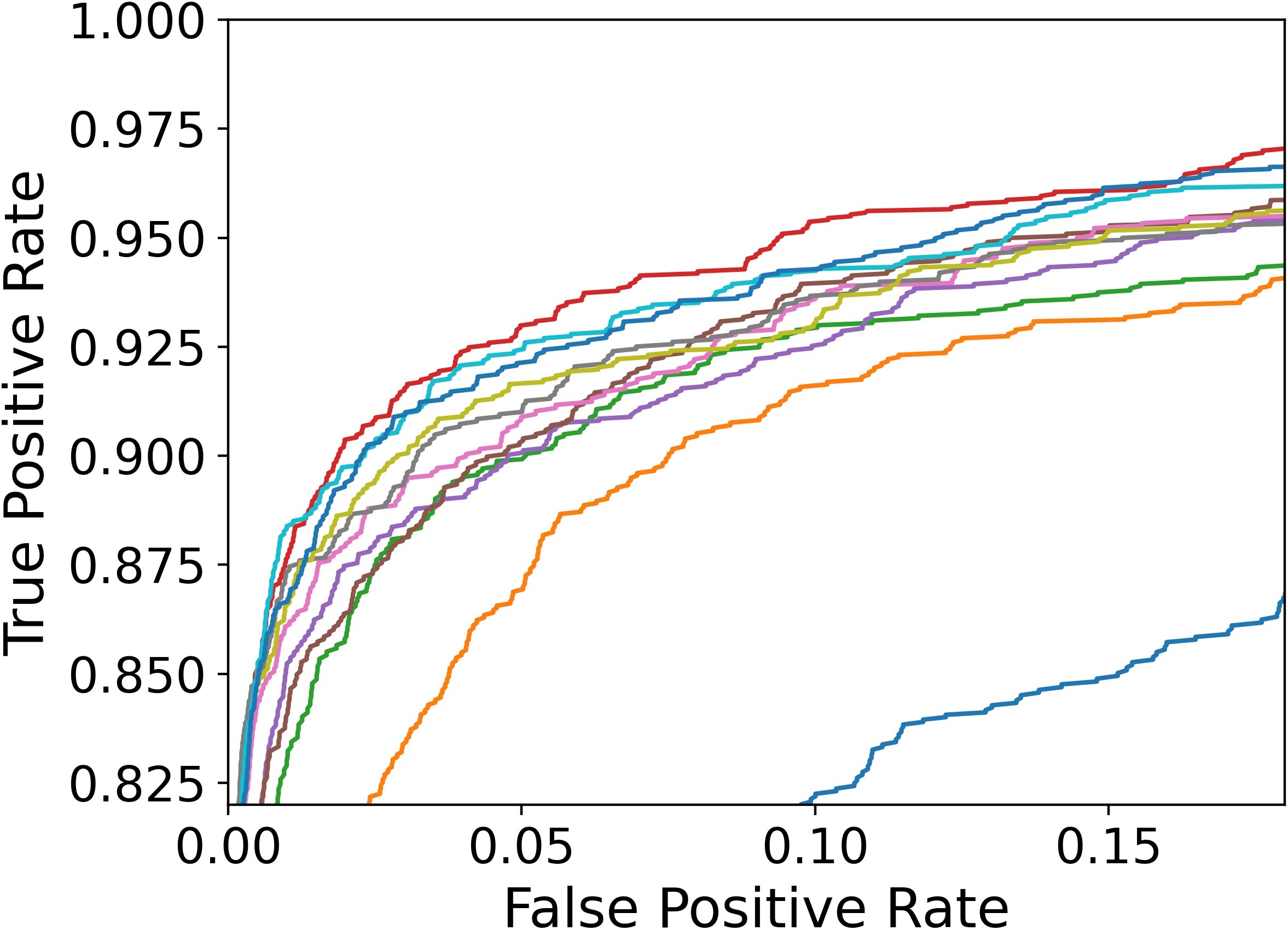}
  \captionof{figure}{ROC curves of TL models and Standard CNN (we magnify the top left part of the curves in the right plot).}
  \label{fig:roccurve}
\end{minipage}%
\hspace{0.5em}
\begin{minipage}{.42\textwidth}
  \centering
  \includegraphics[width=0.98\textwidth]{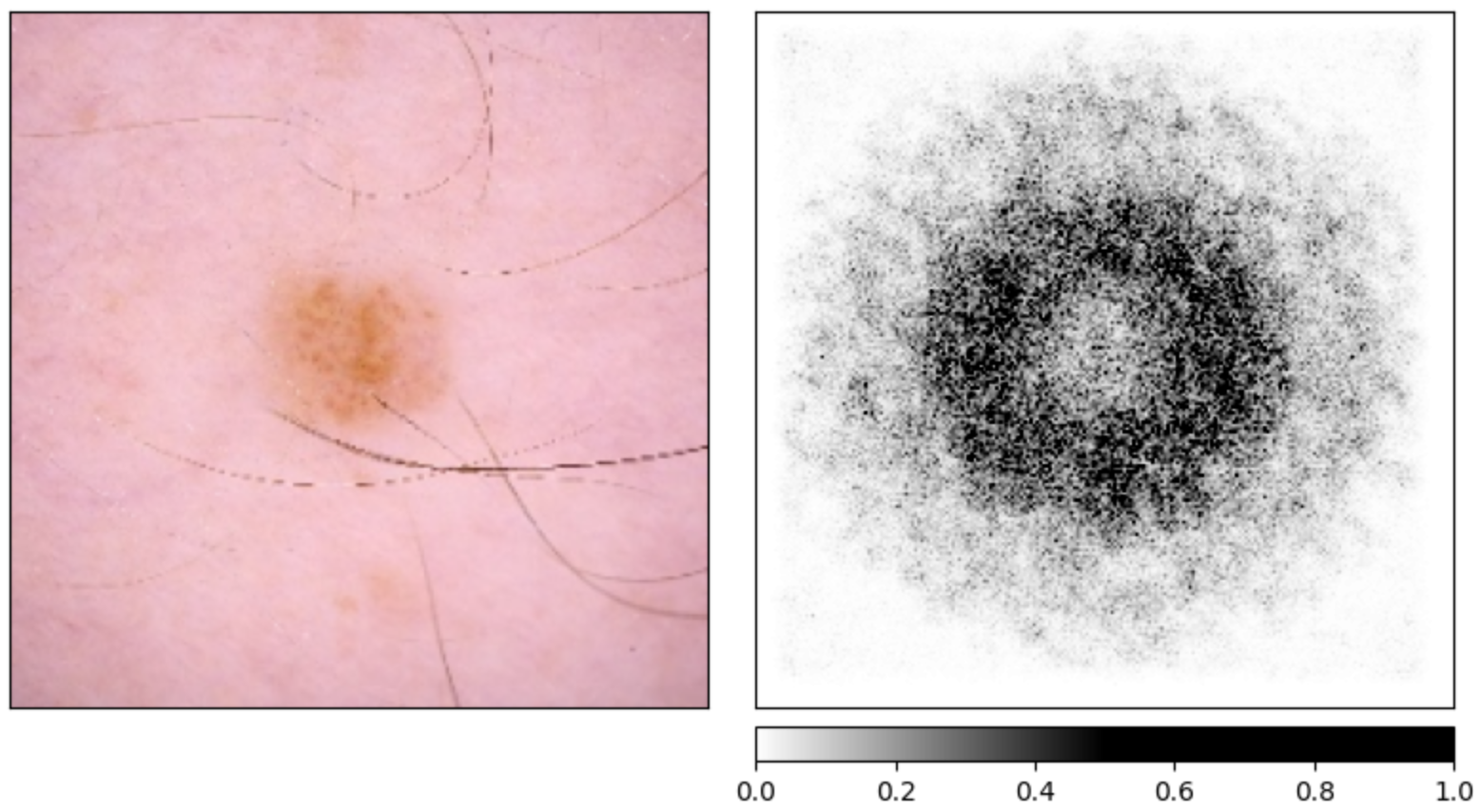}
  \captionof{figure}{A test set image and corresponding integrated gradient attributions for the standard CNN model.}
  \label{fig:visualisation}
\end{minipage}
\end{figure}

Table~\ref{tab:results} shows the test performance of all the models ((a) without and (b) with transfer learning). Eight of the ten CNN models performed significantly better with TL across all 4 metrics, and none of the models are significantly harmed by TL on any metric, demonstrating a clear benefit of transfer learning for melanoma detection. The best performing models are EfficientNet~\cite{effNet} and MnasNet~\cite{MnasNet}, which significantly outperform dermatologists~\cite{Phillips2019} on AUROC. From the ROC curves shown in Figure~\ref{fig:roccurve}, we see that the EfficientNet can achieve a true positive rate of 0.95 while only conceding 0.1 to the false positive rate. 

\begin{table*}[h]
  \caption{\small Performance of the models averaged over 10 independent training runs. Tables (a) and (b) show the performance without and with transfer learning respectively.  The error margins are 95\% confidence intervals (CIs). We report the accuracy, area under the receiver operating characteristic curve (AUROC), area under the precision recall curve (AUPRC) and the F1 Score. Within each table, the results are ordered from least to best performance. In table (b), if the result is statistically better than the model without transfer learning in a one-tailed t-test ($p < 0.05^{*}$ and $p < 0.001^{**}$), then it is indicated with stars. Results that significantly outperform general practitioners and dermatologists on AUROC (determined by a recent meta-analysis$\dag$~\cite{Phillips2019}) are indicated in \textBF{\textcolor{ForestGreen}{green}} and \textBF{\textcolor{blue}{blue}} respectively ($p < 0.05$).}
  \label{tab:results}
  \centering
  \small
  \begin{minipage}{0.04\textwidth}
  \centering
  \subcaption{}
  \end{minipage}
  \begin{minipage}{0.95\textwidth}
  \centering
  \begin{tabular}{p{2.8cm}p{1.8cm}p{1.8cm}p{1.8cm}p{1.8cm}}
    \toprule
        \textBF{Model} & \textBF{Accuracy} & \textBF{AUROC} & \textBF{AUPRC} & \textBF{F1 Score} \\
    \midrule
        Standard CNN & 0.914$\pm$0.004 & 0.759$\pm$0.030 & 0.484$\pm$0.026 & 0.633$\pm$0.035 \\
        General Practitioners$\dag$ & - & 0.83$\pm$0.03 & - & - \\
        VGG~\cite{vgg} & 0.943$\pm$0.004 & 0.832$\pm$0.018 & 0.643$\pm$0.025 & 0.765$\pm$0.023 \\
        SqueezeNet~\cite{SqueezeNet} & 0.949$\pm$0.003 & 0.860$\pm$0.014 & 0.687$\pm$0.011 & 0.801$\pm$0.008 \\
        ResNeXt~\cite{ResNeXt} & 0.952$\pm$0.009 & \textBF{\textcolor{ForestGreen}{0.878$\pm$0.022}} & 0.712$\pm$0.035 & 0.818$\pm$0.023 \\
        DenseNet~\cite{DenseNet} & 0.957$\pm$0.003 & 0.859$\pm$0.015 & 0.733$\pm$0.021 & 0.824$\pm$0.018 \\
        GoogleNet~\cite{googleNet} & 0.957$\pm$0.004 & 0.861$\pm$0.018 & 0.732$\pm$0.024 & 0.824$\pm$0.022 \\
        ResNet-50~\cite{ResNet} & 0.959$\pm$0.003 & \textBF{\textcolor{ForestGreen}{0.869$\pm$0.016}} & 0.744$\pm$0.018 & 0.835$\pm$0.016 \\
        MobileNet~\cite{mobilenetv2} & 0.963$\pm$0.003 & \textBF{\textcolor{ForestGreen}{0.889$\pm$0.013}} & 0.769$\pm$0.019 & 0.856$\pm$0.014 \\
        MnasNet~\cite{MnasNet} & 0.963$\pm$0.008 & \textBF{\textcolor{ForestGreen}{0.900$\pm$0.010}} & 0.771$\pm$0.039 & 0.859$\pm$0.023 \\
        ShuffleNet~\cite{shufflenet} & 0.965$\pm$0.004 & \textBF{\textcolor{ForestGreen}{0.892$\pm$0.016}} & 0.777$\pm$0.025 & 0.861$\pm$0.018 \\
        EfficientNet~\cite{effNet} & 0.967$\pm$0.002 & \textBF{\textcolor{ForestGreen}{0.900$\pm$0.009}} & 0.794$\pm$0.013 & 0.872$\pm$0.010 \\
        Dermatologists$\dag$ & - & \textBF{\textcolor{ForestGreen}{0.91$\pm$0.02}} & - & - \\
    \bottomrule
    \end{tabular}
    \end{minipage}
    \begin{minipage}{0.04\textwidth}
    \centering
    \subcaption{}
    \label{tab:dynamic}
    \end{minipage}
    \begin{minipage}{0.95\textwidth}
    \centering
    \begin{tabular}{p{2.8cm}p{1.8cm}p{1.8cm}p{1.8cm}p{1.8cm}}
    \midrule
        General Practitioners$\dag$ & - & 0.83$\pm$0.03 & - & - \\
        VGG~\cite{vgg} & 0.959$\pm$0.003$^{**}$ & \textBF{\textcolor{ForestGreen}{0.874$\pm$0.013}}$^{**}$ & 0.740$\pm$0.016$^{**}$ & 0.835$\pm$0.013$^{**}$ \\
        ResNet-50~\cite{ResNet} & 0.962$\pm$0.004 & \textBF{\textcolor{ForestGreen}{0.880$\pm$0.014}} & 0.763$\pm$0.022 & 0.849$\pm$0.017 \\
        ShuffleNet~\cite{shufflenet} & 0.963$\pm$0.006 & \textBF{\textcolor{ForestGreen}{0.896$\pm$0.024}} & 0.769$\pm$0.040 & 0.857$\pm$0.028 \\
        SqueezeNet~\cite{SqueezeNet} & 0.963$\pm$0.004$^{**}$ & \textBF{\textcolor{ForestGreen}{0.902$\pm$0.015}}$^{**}$ & 0.771$\pm$0.020$^{**}$ & 0.861$\pm$0.015$^{**}$ \\
        DenseNet~\cite{DenseNet} & 0.966$\pm$0.003$^{**}$ & \textBF{\textcolor{ForestGreen}{0.904$\pm$0.011}}$^{**}$ & 0.786$\pm$0.018$^{**}$ & 0.870$\pm$0.011$^{**}$ \\
        Dermatologists$\dag$ & - & \textBF{\textcolor{ForestGreen}{0.91$\pm$0.02}} & - & - \\
        MobileNet~\cite{mobilenetv2} & 0.969$\pm$0.002$^{**}$ & \textBF{\textcolor{ForestGreen}{0.916$\pm$0.007}}$^{**}$ & 0.806$\pm$0.015$^{**}$ & 0.884$\pm$0.009$^{**}$ \\
        ResNeXt~\cite{ResNeXt} & 0.971$\pm$0.001$^{**}$ & \textBF{\textcolor{ForestGreen}{0.918$\pm$0.006}}$^{**}$ & 0.819$\pm$0.009$^{**}$ & 0.891$\pm$0.005$^{**}$ \\
        GoogleNet~\cite{googleNet} & 0.973$\pm$0.002$^{**}$ & \textBF{\textcolor{ForestGreen}{0.921$\pm$0.006}}$^{**}$ & 0.831$\pm$0.013$^{**}$ & 0.898$\pm$0.008$^{**}$ \\
        MnasNet~\cite{MnasNet} & 0.974$\pm$0.002$^{*}$ & \textBF{\textcolor{blue}{0.928$\pm$0.005}}$^{**}$ & 0.832$\pm$0.013$^{**}$ & 0.901$\pm$0.007$^{**}$ \\
        EfficientNet~\cite{effNet} & 0.975$\pm$0.002$^{**}$ & \textBF{\textcolor{blue}{0.931$\pm$0.005}}$^{**}$ & 0.840$\pm$0.010$^{**}$ & 0.906$\pm$0.006$^{**}$ \\
    \bottomrule
    \end{tabular}
    \end{minipage}
\end{table*}

\subsection{Visualisation}
We used the integrated gradients method~\citep{integratedgradients} to calculate feature attributions. This method computes the importance scores $\phi_i^{IG}$ by accumulating gradients interpolated between a baseline $\textbf{b}$ input (intended to represent the absence of data, in our case this is a black image) and the current input $\textbf{x}$.
\begin{equation}
    \phi_i^{IG}(\psi, \textbf{x}, \textbf{b}) = \overbrace{(\textbf{x}_i - \textbf{b}_i)}^{\text{diff. from baseline}} \times \int_{\alpha=0}^{1}\overbrace{\frac{\delta \psi (\textbf{b} + \alpha(\textbf{x} - \textbf{b}))}{\delta \textbf{x}_i}}^{\text{acc. local grad.}}d \alpha
\end{equation}
The CNN model is represented as $\psi$\footnote{The background and intuition behind the method is explained clearly in \citet{sturmfels2020visualizing}.}. We observed that the models tend to focus primarily on the edges of the skin lesions (Figure~\ref{fig:visualisation}). This aligns with our expectation, since uneven or notched edges are common in melanoma~\cite{McCourt2014}. Secondary to the edges, there is some importance to the lesion itself and surrounding skin. This is significant because melanomas can also show uneven texture or colour~\cite{McCourt2014}.

\section{Conclusion}
We have conducted an extensive investigation of transfer learning for the task of melanoma detection from photographs. We have demonstrated the benefit of transfer learning with ImageNet pre-training~\cite{imagenet} for melanoma detection on the ISIC 2020 dataset~\cite{dataset1}. Furthermore, we show that the best performing neural networks are EfficientNet and MnasNet, which are capable of outperforming dermatologists when distinguishing melanoma from benign skin lesions. In particular, we note that these networks have been specifically designed for mobile devices~\cite{effNet,MnasNet}. This may be important when it comes to data privacy and medical data regulations (as the classification can be performed locally on the user's personal device).

In future work, we aim to extend the binary classification task to classify other skin lesions such as benign keratosis, basal cell carcinoma, actinic keratosis, vascular lesions and dermatofibroma. Secondly, we would like to extend our interpretability study such that we can visualise the learnt features in the intermediate layers of the models. To do this we can leverage the approach of~\citet{45507} whereby we obtain inputs designed to maximise the activation of hidden layers of the network. This will provide further insights as to why certain models outperform others in melanoma detection. Finally, we can validate the diagnostic technology in the community with an implementation study.


\section{Broader Impact}
The automated diagnosis technology could be used to screen, triage, refer and follow-up patients in the community. It also has potential to reach patients who would not normally have access to dermatologists e.g. in remote areas or the developing world. The high AUROC of EfficientNet (high true positive rate coinciding with a low false positive rate) would make it well-suited to this purpose. Such a system could significantly reduce the cost and resources needed to screen and treat as it reduces the pool of patients needing to see the dermatologist.

\section*{Acknowledgements}
The authors would like to thank Horizon Academic for facilitating this research.

\bibliography{references}

\newpage
\appendix

\section{Transfer Learning Architectures}
\label{transfermodels}
\paragraph{VGG}
VGG~\cite{vgg} is an advancement of a previous deep neural network, AlexNet~\cite{alex}. The model uses small receptive fields of 3x3 with five max-pooling layers. In our paper, we use VGG16.
\paragraph{GoogleNet}
GoogleNet~\cite{googleNet} was developed in 2014 to solve the problem of overfitting by building an Inception Module, using filters of multiple sizes. Three filter sizes of 1x1, 3x3, and 5x5 are simultaneously used; whereby the 1x1 convolution is used to shrink the dimensions of the model. The GoogleNet architecture consists of 9 Inception Modules, with each module connected to an average pooling layer.
\paragraph{ResNet}
ResNet \cite{ResNet} short for ``Residual Network'', is a deep learning model developed in 2015 and was the winner of the ImageNet Competition \cite{imagenet}. In our research, we use ResNet50, a variant of the ResNet Model. The model consists of 48 Convolutional layers, 1 Max Pooling and 1 Average Pooling layer. ResNet addresses the vanishing-exploding gradients by leveraging skip connections for identity mapping, simplifying the network.
\paragraph{SqueezeNet}
SqueezeNet~\cite{SqueezeNet} uses fewer parameters while preserving similar performance to AlexNet~\cite{alex}. There are several architectural features worth noting: the use of 1x1 convolution filters, decreased number of input channels, and down-sampling later in the network.
\paragraph{DenseNet}
DenseNet \cite{DenseNet} is similar to the architecture of ResNet but with ``DenseBlocks''. Each DenseBlock consists of a convolution layer, pooling layer, batch normalisation, and non-linear activation layer.
\paragraph{ResNeXt}
Built on the Residual Network and VGG, ResNeXt \cite{ResNeXt} uses a similar split-transform-merge strategy with an additional cardinality dimension (size of a set of transformations). The models borrows the repeating layers strategy from VGG and ResNet and according to the researchers, has better performance than ResNet \cite{ResNet} but with only 50\% complexity.
\paragraph{MobileNet}
MobileNet \cite{mobilenetv2} was developed for devices with smaller computational power such as smartphones. Unlike bigger deep learning networks such as VGG, MobileNet uses depthwise separable convolution, performing convolution on the input channels separately and then by pointwise convolution. This way low latency models can be developed, which are applicable to mobile devices.
\paragraph{ShuffleNet}
ShuffleNet \cite{shufflenet} was also designed for mobile devices with small computational power. The model uses 1x1 convolution and channel shuffle, designed specifically for small networks. ShuffleNet has efficient computation while obtaining an accuracy similar to and thirteen times faster than AlexNet~\cite{alex}.
\paragraph{MnasNet}
MnasNet \cite{MnasNet} is another mobile network designed for efficient performance using a multi-objective neural architecture search approach that considers accuracy and latency. The network also uses a hierarchical search space, achieving speeds faster than MobileNet~\cite{mobilenetv2}. 
\paragraph{EfficientNet}
EfficientNet~\cite{effNet} is a recent mobile network developed in 2018, which applies a compound coefficient for improved accuracy. Rather than scaling up the CNN model by an arbitrary amount, the authors use a grid search to find correlation in scaling based on the AutoML neural architecture search.

\section{Hyperparameter Optimisation}
\label{hyperparametertuning}
For the standard CNN model, values in a range were tested for dropout, batch size, kernel size, learning rate, number of layers, pool size, and number of convolution filters using random search, which is a more efficient method of hyperparameter optimisation than grid or manual search \cite{randomsearch}. The search ranges and final values are shown in Table ~\ref{hyperparamtable}.

\begin{table}[h!]
\centering
\begin{tabular}{lllll}
\toprule
\textbf{Hyperparameter} & \textbf{Value} & \textbf{Lower} & \textbf{Upper} & \textbf{Scale} \\ \midrule
Dropout & 0.4 & 0.0 & 0.5 & Linear \\ 
Batch Size & 8 & 4 & 512 & log$_{2}$ \\
Kernel Size & 4 & 2 & 5 & Linear \\ 
Learning Rate & 0.00977 & 0.001 & 0.01 & log$_{10}$ \\ 
Number of Layers & 5 & 5 & 10 & Linear \\ 
Pool Size & 3 & 3 & 4 & Linear\\ 
Convolution Filters & 11 & 6 & 12 & Linear\\
\bottomrule
\end{tabular}
\vspace{0.5em}
\caption{Hyperparameter search ranges and final values.}
\label{hyperparamtable}
\end{table}

\section{Implementation}
\label{implementation}
All deep learning methods were implemented in PyTorch \citep{NEURIPS2019_9015} and were optimised using Adam \citep{KingmaB14}. The models were trained using Tesla P100 GPUs. Each model was trained over 10 independent training runs with early stopping \cite{earlystopping} for a maximum of 15 epochs (the patience constant was set to 3). We used step decay in the learning rate (the decay was set to 0.4 with a learning patience of 1).

\section{Additional Tables and Figures}

\begin{table}[h]
\centering
\begin{tabular}{lll}
\toprule
\textbf{Location}        & \textbf{Normal} & \textbf{Melanoma} \\
\midrule
Torso           & 17106 & 257 \\
Lower extremity & 8293 & 124 \\
Upper extremity & 4872 & 111 \\
Head/neck       & 1781 & 74 \\
Palms/soles & 370 & 5 \\
Oral/genital & 120 & 4 \\
\bottomrule
\end{tabular}
\vspace{0.5em}
\caption{Distribution of normal and melanoma skin lesion samples based on location of body.}
\label{image_distribution}
\end{table}

\end{document}